\documentclass[conference]{IEEEtran}
\usepackage{amsfonts}

\usepackage{url,graphicx,graphics,times,amsmath,xcolor} 
\usepackage{algorithm}
\usepackage{algorithmic}

\usepackage[american]{babel}
\usepackage[utf8]{inputenc}

\IEEEoverridecommandlockouts   

\newcommand{\cf}{\textit{cf.}}

\newcommand{\etal}{\textit{et al.}}
\newcommand{\E}{\mathbb E}

\begin{document}

\title{\ \\ \LARGE\bf 
  Bandits attack function optimization
  \thanks{Philippe Preux is with Université de Lille, LIFL (UMR 8022 CNRS), and INRIA, France. Rémi Munos and Michal Valko are with INRIA, and LIFL, Lille, France. (email: \{philippe.preux,remi.munos,michal.valko\}@inria.fr).}}

\author{Philippe Preux and Rémi Munos and Michal Valko}


\maketitle

\begin{abstract}
  We consider function optimization as a sequential decision making problem under budget constraint. 
  This constraint limits the number of  objective function evaluations allowed during
  the optimization. We consider an algorithm inspired by a continuous version of a multi-armed bandit problem 
  which attacks this optimization problem by solving the tradeoff between exploration (initial quasi-uniform search of the domain) and exploitation (local optimization around the potentially global maxima).  
  We introduce the so-called Simultaneous Optimistic Optimization (SOO), a deterministic algorithm that works by
  domain partitioning. The benefit of such approach
  are the guarantees on the returned solution and the numerical efficiency of the algorithm. 
  We present this machine learning approach to optimization, and
provide the empirical assessment of SOO on the CEC'2014 competition on single
objective real-parameter numerical optimization test-suite.
\end{abstract}


\section{Introduction}

\PARstart{F}{unction} optimization has been due to its generality, a recurring
topic for centuries. In this work, we assume that the function is available only
through a black-box: one provides a point in space to the black-box which
returns the value of the function at that point. It is customary to make
assumptions about the objective function (the function being optimized), whether
it is continuity, smoothness, or even differentiability. These assumptions are
not necessarily met by the real-world objective functions and commonly they
are impossible to verify. Moreover, the objective function may be
nondeterministic, returning different values along time for the same point. An
important practical aspect is that the evaluation of the objective function at a
given point always takes
some resource (computational time, energy, bandwidth,
economical, \dots) and it is suitable to optimize a
trade-off between the quality of the optimum being found and the amount of
resources that have been used to achieve the result.


Most of real-world functions to optimize are multi-modal. One is then
interested in finding the ``best point'', a \textit{global optimum}, or one
among them if there are many equally optimal points. Basically, we
distinguish \textit{global} or \textit{local} numerical
optimization algorithms. To avoid the ambiguity about how
the term ``global'', we mean that the algorithm is guaranteed to
return a global optimum given enough resources. Therefore, we
first require an asymptotic guarantee of optimality. More interestingly, we aim
for the approach with a a finite-time guarantee, where we provably evaluate
the quality of the returned solution as a function of
the evaluation budget. This comes in contrast with local algorithms that can
only be proven to return a local optimum.

With no prior knowledge on the function smoothness such a global
optimization algorithm would perform iteratively the following action: based on
already evaluated points and their values, choose the next point to evaluate.
Deciding on an action based on past actions and their consequences is known as
\textit{sequential decision making}. Since the objective function is
unknown, this problem is \textit{under uncertainty}. There is a vast body of
literature on this problem, coming from the multitude of scientific domains such
as machine learning, statistics, and operation research.
In this paper, we wish to build on the existing body of work and bring
a principled approach to function optimization. Hence, we introduce the
necessary background to derive an algorithm which is fairly simple but
performs well in practice and which properties can be studied theoretically.
We regard the main contribution of this paper the idea
that function optimization is a sequential decision making 
problem which opens new ways of approaching it.

In the rest of this paper, we first introduce the framework of sequential
decision making under uncertainty, and current state of the art concepts that
will be used to design our function optimization algorithm. Next, we introduce
our algorithm acronym-ed by SOO and discuss some of its properties. In
section~\ref{sec:expe}, we perform the experimental assessment of SOO based on
the \textit{CEC'2014 competition on real parameter numerical optimization} test
suite. We think that the results show that SOO is a really serious alternative
for function optimization. Based on these results, we discuss the strengths and
weaknesses of SOO and its application.

\section{Function optimization and sequential decision making under uncertainty}

As the background of this work stems from the ``sequential decision making
problem under uncertainty'',  we first describe the necessary basics. For
the sake of adequacy we restrict ourselves to the relevant
parts.

Imagine an agent interacting with its environment; assuming that time
is discrete. The agent perceives its environment in some way and finds itself
in a \textit{state}; the agent possesses a certain repertoire of actions; at
each time step -- since is a \textit{sequential} process -- the agent decides
which action to perform based on its previous history, in order to reach a
certain goal. The goal is formalized by an objective function which is
not fully known to the agent: the agent somehow probes the objective function
through its actions but not directly through the objective function. When the
agent performs an action, it receives a consequence, also known as a return or
a \textit{reward}; this consequence reflects the value of the action for the
objective function.

Let us formalize the setting above: an agent faces a set of $K$ actions.
Each action $a_k$ is associated to a certain distribution $\nu_k$, whose expectation 
is denoted by $\mu_k=E[\nu_k]$. At any round $t$, the agent executes an action $k_t\in\{1,\dots K\}$
and receives a noisy sample $r_t$ -called reward- drawn (independently) from the distribution $\nu_{k_t}$ 
(thus we have $\E[r_t] = \mu_{k_t}$).
The goal of the agent is to find the strategy that maximizes the sum of collected rewards in expectation. 
This problem is known as the ``multi-armed bandit
problem'', introduced in 1933 by W.R. Thompson \cite{thompson1933} and
independently in 1952 by H. Robbins \cite{robbins1952}. This is also known as
the exploration vs.\@ exploitation dilemma. A related question is: how can the agent identify as fast as possible
the best action, that is the one which is the most rewarding on average
($k^*=\arg\max_k \mu_k$). This has been called the simple regret problem or best arm identification in multi-armed bandit settings \cite{BMS09,colt2010b}.

This problem has witnessed remarkable advances within the last 15 years thanks
to the introduction of the notion of \textit{upper confidence bounds} \cite{ucb}. 
We illustrate the approach with a simple example: Suppose the agent
faces $K=2$ possible actions, $a_1$ and $a_2$; At time $t=15$, the action
$a_1$ has been executed by the agent 5 times, with an average reward of $0.3$,
while $a_2$ has been executed 10 times resulting in the same average rewards of $0.4$: 
at the next time step, should the agent perform $a_1$ or $a_2$? There are
many strategies to cope with this situation, well-known in the EC community
($\varepsilon$-greedy, softmax, Boltzmann-Gibbs, ...). The ``Upper-Confidence Bound'' (UCB) 
approach that we consider here follows the so-called ``optimism in the face of uncertainty'' principle, 
which consists in selecting the action that has the highest rewards in all possible
environments that are reasonably compatible with the observations (i.e.~rewards
observed so far). Here, action $a_2$ seems better than $a_1$ in terms of average
empirical rewards. However action $a_1$ has been chosen less often than $a_2$, thus the
uncertainty over its true mean $\mu_1$ is higher than for $\mu_2$, thus the set of possible
values of $\mu_1$ given the observed rewards is potentially larger than for action $2$. Here
exploitation would mean selecting action $a_2$ since its empirical mean is 
higher than $a_1$, but exploration would imply selecting $a_1$ in order to get 
additional information about action $a_1$ and thus reduce the uncertainty over $\mu_1$. The UCB algorithm
solves this exploration-exploitation tradeoff by defining an UCB on the mean of
each action based on the rewards observed so far and the number of times each action have been chosen.
Then the action with highest UCB is selected.

To be more specific, assume that each action $k$ has been executed $n_{k,t}$
times up to time $t$ and the average reward observed on this action is
$\hat{\mu}_{k,t}$. Then, the UCB strategy selects the action $k_t$ that
maximizes the following UCBs (where the ``2'' that appears in the equation is actually a tunable
constant):
\begin{eqnarray}
  k_t = \arg\max_{1\leq k\leq K} \Big( \hat{\mu}_{k,t} + \sqrt{\frac{2 \log t}{n_{k,t}}}\Big)
\end{eqnarray}

One can show that the choice of UCB leads to a expected cumulative performance (sum of obtained rewards) up to time $n$ 
which is almost as high as the optimal value (which would be $n\max_k \mu_k$, i.e.~$n$ times the mean of the best action) if we knew the action distributions.
Also, following the UCB algorithm, one may prove that the number of time steps at which
suboptimal actions are executed grows only logarithmically with time, that is
the exploration cost grows logarithmically with time, which is optimal in this setting
(there exists lower-bounds saying that one cannot expect to make less than a logarithmic number
of mistakes, see \cite{lr1985}).

In the related (simple regret or best-arm identification) setting, 
one is given a fixed budget $n$ of action executions, and the objective is 
to explore as efficiently as possible the set of actions during $n$ rounds and make a final 
recommendation of what is the best action. This formulation is closer in spirit to the 
problem of function optimization under budget constraint but share strong links 
(in terms of exploration-exploitation tradeoff) with the previous
problem of maximizing the sum of rewards, see \cite{BMS09,colt2010b}.

\section{Simultaneous Optimistic Optimization}
\label{sec:soo}

\subsection{The algorithm}

The application of the previous framework to function approximation over a continuous domain is interesting. 
The set of possible actions is the domain of definition of the
function to optimize and the reward subsequent to an action execution is the
value of the objective function at the point associated to the action.

At first, the reader may raise serious concern about the fact that the set of possible actions is hence uncountable. It turns out that this problem can be overcome, both in theory and in practice, by relying on some smoothness assumption of the objective function. This setting falls in the category of ``structured bandit problem''.
Of course, one would like to set minimal assumption on the target function. However the optimization problem is known to suffer from the curse of dimension, even in finite spaces. Also the no-free-lunch theorem somehow tells us that for any clever algorithm, there exists a problem on which the clever algorithm performs poorer than a stupid algorithm (say an algorithm that just explores uniformly at random). However if we restrict the classes of functions of interest, then it is possible to define algorithms that would be good on such classes. Now, ideally we would like to assume some smoothness on the objective function, but without having to tune the algorithm according to the specific smoothness of the function (which is usually unknown).

This is the reason we rely on an algorithm called ``Simultaneous Optimistic Optimization'' (SOO), which has been introduced by Munos in 2011 \cite{soo:nips2011d}. SOO is a deterministic algorithm, meant to optimize deterministic functions which are assumed to be locally smooth near their global optima (in a specific sense) but where the actual smoothness does not need to be known. The actual smoothness is not needed by the algorithm, but the performance of the algorithm will depend on this smoothness.

\begin{figure}
 \begin{center}
 \includegraphics[height=12cm]{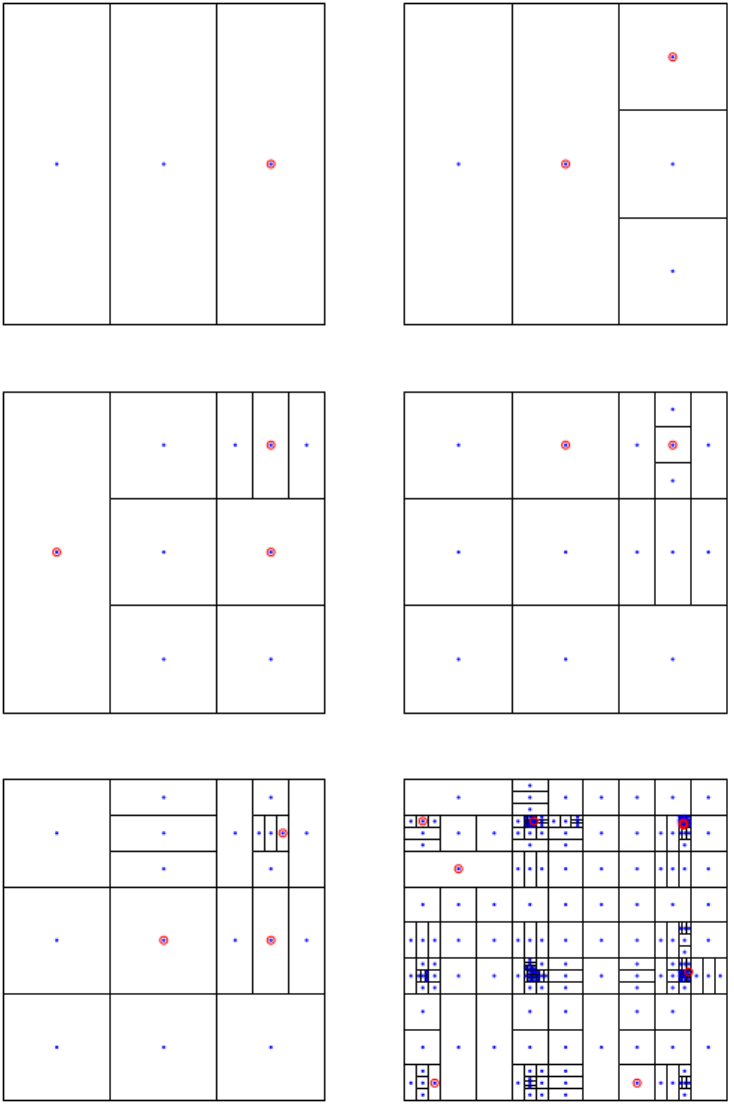}
  \caption{An illustration of SOO at work during a function approximation.
Assuming a square bi-dimensional domain of search for the optimum and $S=3$, the
first steps splits the initial cell into $S$ sub-cells (top left); the split
occurs along one dimension. The next step splits one of those sub-cells (top
right). Steps 1 to 5 are represented from top to bottom. Bottom right is the
partition obtained after 150 steps on a certain function. At each step, the cell
being split as its center circled in red.}
  \label{fig:cellsplit}  
 \end{center}
\end{figure}

SOO works by partitioning the domain of definition of the objective function. At first, the whole domain of search is one cell. This cell (as well as all forthcoming cells) is represented by the point at its center, and the value of the objective function for this point. The first iteration consists in splitting this cell into $S$ sub-cells; the cut occurs along one dimension. The center point of each sub-cell is evaluated. Assuming we minimize the objective function, SOO chooses to split the cell with the smallest value at its center; the split then occurs along an other dimension (with regards to the initial split), and the center point of each of these sub-cells is evaluated. By so doing, SOO builds a tree of cells, the root being the entire search domain. Then, at each iteration, SOO considers each depth in the tree and for each depth, select the cell with lowest value at this depth and splits this cell {\em only} if its value is lower than all previously selected cells of lower depths. The iterations go on until the budget of function evaluations is exhausted. The point with lowest evaluation is then returned.
Fig.\@ \ref{fig:cellsplit} illustrates the procedure.
Taking $S$ odd, the cell being split shares its center with one of the sub-cells so that each cell split involves $S-1$ objective function evaluations.
That way, SOO performs adaptive partitioning, focusing on the area of the domain
which is the most promising at each iteration.



\subsection{Properties of SOO}

\begin{itemize}
  \item SOO is very simple, hence very easy to implement in a very efficient manner.
  \item SOO is a rank-based algorithm (the actual values of $f$ do not matter, but only their pairwise comparisons)
  \item SOO does not rely on the knowledge of the smoothness of the objective function.
  \item SOO is consistent: it converges asymptotically towards the global optimum
  \item after $t$ time steps, the expected difference between the value of the global optimum of the objective function $f^*=\min_{x\in X}f(x)$ and the value of the best point $x_t$ returned by SOO is:

\begin{eqnarray}
  f(x_t) - f^* \mbox{ is decreasing in } O(t^{-1/d}),
\end{eqnarray}
where $d$ is the near-optimality dimension of $f$ (defined precisely in \cite{soo:nips2011d}), which provides a measure of the quantity of near-optimal points near the global optimum of $f$. Note that exponential rate $O(e^{-ct})$ can also be achieved in the non-trivial case $d=0$.  This is a nice result and SOO is unique as being a global optimization algorithm for which such a finite-time performance result have been formally proven under such weak assumption on the objective function.
\end{itemize}

\subsection{Related algorithms}

SOO is a global function optimization algorithm; it may be related to many other algorithms, in particular with ES, CMA-ES, and other evolutionary algorithms. However, SOO is deterministic. Actually, SOO is very closely related to the algorithm DiRect introduced by Jones \etal{} in 1991 \cite{direct}. DiRect stands for ``Dividing Rectangles''. DiRect assumes the objective function to be globally Lipschitz (with unknown Lipschitz constant) whereas SOO assumes a much weaker assumption on $f$ which is only local (around the global optimum). In addition finite-time performance guarantees are obtained for SOO whereas DiRect only enjoys an (asymptotic) consistency result.

\section{Experimental evaluation of SOO}
\label{sec:expe}

The design of SOO stems from research in bandit theory which aims at proposing algorithms which are amenable to a theoretical analysis of their performance. Actually, to the best of our knowledge, SOO is the only existing global function optimization algorithm for which the discrepancy between the best found point after $N$ function evaluations and the optimum is known under very weak assumptions. That being said on the theoretical basis, we were eager to assess its performance in practice. \textit{A priori}, our expectation was that such a partitioning algorithm would be able to deal with functions up to 5 dimensions, and would perform rather badly beyond.

Hence the results of the experimental assessment discussed in this section come as a surprise. Though SOO does have difficulties on some functions in 10D, SOO is still able to perform remarkably well in 100D on some other functions.

\subsection{Set-up}

The setup of the experimental evaluation of SOO is that of the CEC'2014 competition on single objective real-parameter numerical optimization (see \cite{cec2014CompetitionA}). 30 objective functions should be optimized in 10, 30, 50, and 100 dimensions with a given budget of function evaluations. The budget is proportional to the dimension: $10^4 \times$ the dimension, hence $10^5$, $3\ 10^5$, $5\ 10^5$, and $10^6$ function evaluations. Unless otherwise stated, all results are provided within this setup.
Objective functions are numbered from 1 to 30, and the optimum value is 100 times the number of the function (that is, 100 for function 1, up to 3000 for function 30).
Functions 1 to 16 may be said to be atomic, whereas functions 17 to 22 are sums of 3, 4 or 5 atomic functions, and functions 23 to 30 are a weighted sum of 3, 4 or 5 atomic functions.

\subsection{SOO parameters}

SOO has just a few parameters:

\begin{itemize}
  \item $S$: the number of cells resulting from the split of a cell.
  \item the maximum depth of the tree (as a function of $N$)
  \item the direction of split of cells.
\end{itemize}

We did a little exploration of parameter values and ended using default values since the changes in performance were rather haphazardous, and not so significant.

The selected cell is thus split into $S=3$ sub-cells. The maximum depth of the tree is $\sqrt{(\log{N})^3}$ (set taking inspiration from \cf{} \cite{icml2013b}). The dimensions being numbered, the direction of a split is merely the next dimension with regards to the previous one.

\subsection{Results}

Regarding notations, for an objective function $f$, we note $f(x^*)$ the value of the best point found by SOO, and $f^*$ the value of the optimum.

\subsubsection{Running time}

All experiments were developed and executed on a Lenovo Thinkpad X220 based on an Intel Core i7-2640M CPU, 2.80GHz, 8Gb of main memory. The computer runs Ubuntu 12.04. All software is compiled with \verb|gcc| version 4.6.3 using aggressive code optimization options.

On function 18, the running time (wall clock) of the execution of SOO is about 3 minutes in 10D, 12 minutes in 30D, 30 minutes in 50D, 110 minutes in 100D. The computation time of the objective function itself amounts to less of a few seconds (depending on the dimension); so, this time is mostly used by SOO.

\subsubsection{Dependence on dimension}

We run SOO on all 30 functions defined in 10, 30, 50, and 100 dimensions. 
We use the same number of function evaluations regardless of the dimension of the problem (which is not the competition protocol).
The results are given in table \ref{table:dim}.
Obviously, we observe a degradation of results when the dimension increases. However, for a very significant subset of functions, the degradation is far from being severe (to our own surprise). Indeed, for functions 5, 12, 13, 14, 23, 28, 29, and 30, the degradation between 10 D and 100 D counts as less as $0.5$\%; for functions 16, 24, 25, 26, and 27, the best found point is still very close to the optimum, their value being less than 10\% larger that the optimum.
It is also true that the performance are terrible on some of the objective functions, such as functions 1 and 2, and to a lesser extent on functions 3, 17, and 18. Functions 1 and 2 are difficult to optimize because the optimum is located in a very narrow region. Note that functions 17 and 18 combine 3 atomic functions, among which function 1 for 17, and function 2 for 18.

\begin{table*}
  \begin{center}
    \caption{For each function of the CEC'2014 competition defined in dimensions 10, 30, 50, and 100, this table gives the ratio $\frac{f(x^*)}{f^*}$. In each dimension, the number of objective function evaluations is $10^5$ whatever the dimension of the domain: please note that this is NOT CEC'2014 competition setup.}
    \label{table:dim}
    \begin{tabular}{cccccc}
      Function & 10 D & 30 D & 50 D & 100 D \\ \hline
 1&    1139.36000&   502760.00000&   500290.00000&  2302870.00000\\
 2& 1014070.00000& 13003300.00000& 30815100.00000& 73601500.00000\\
 3&       8.73205&       41.86750&       48.21300&      165.01900\\
 4&       1.01269&        1.49726&        2.32909&        5.25297\\
 5&       1.04000&        1.04034&        1.04060&        1.04155\\
 6&       1.00255&        1.01285&        1.05010&        1.15777\\
 7&       1.00103&        1.00500&        1.07859&        1.23042\\
 8&       1.01751&        1.08213&        1.19856&        1.50861\\
 9&       1.02105&        1.10622&        1.25049&        1.71612\\
10&       1.16098&        2.76473&        5.22053&       13.60660\\
11&       1.22269&        3.17396&        6.08450&       15.53040\\
12&       1.00017&        1.00015&        1.00032&        1.00067\\
13&       1.00004&        1.00034&        1.00053&        1.00052\\
14&       1.00011&        1.00089&        1.00070&        1.00010\\
15&       1.00059&        1.02031&        1.19003&        2.85522\\
16&       1.00140&        1.00697&        1.01093&        1.02527\\ \hline
17&     233.67300&     3880.91000&    25648.20000&    44077.70000\\
18&       8.23793&     1945.28000&     7180.64000&   277819.00000\\
19&       1.00137&        1.01835&        1.05569&        1.19766\\
20&       1.12523&       13.64210&       22.18100&       49.81050\\
21&       2.50779&       99.77110&     4411.87000&    42602.70000\\
22&       1.00992&        1.22135&        1.82905&        2.57441\\ \hline
23&       1.08696&        1.08696&        1.08696&        1.08696\\
24&       1.05009&        1.08333&        1.08333&        1.08333\\
25&       1.06362&        1.08000&        1.08000&        1.08000\\
26&       1.03850&        1.07692&        1.07692&        1.07692\\
27&       1.00220&        1.07407&        1.07407&        1.07407\\
28&       1.07143&        1.07143&        1.07143&        1.07143\\
29&       1.06897&        1.06897&        1.06897&        1.06897\\
30&       1.06667&        1.06667&        1.06667&        1.06667\\
    \end{tabular}
  \end{center}
\end{table*}

Following CEC'2014 competition protocol in which the number of function evaluations scales with the dimension ($10^4 \times$ dimension of the problem), the results after the last function evaluation are given in table \ref{table:dim2}.

Thanks to the use of 3 times more function evaluations in 30D (resp.\@ 5 times
more in 50D), there is an average improvement of 9\% (resp.\@ 12\%) of the
minimum, the standard deviation of their improvement being 20\% (resp.\@ 21\%).
The median is $0.3$\% which reveals that most improvement are small (in 50D, the
median of the improvement is larger: 6\%).

\begin{table*}
  \begin{center}
    \caption{For each function of the CEC'2014 competition defined in dimensions 10, 30, 50, and 100, this table gives the ratio $\frac{f(x^*)}{f^*}$. In each dimension, the number of objective function evaluations is $10^4\times$dimension of the problem. This is CEC'2014 competition setup. The results in 10 dimensions are the same as in table \ref{table:dim}; we mention them in the table to ease comparisons.}
    \label{table:dim2}
    \begin{tabular}{cccccc}
      Function & 10 D & 30 D & 50 D \\ \hline
        1 & 1139.36000 & 319606.00000 & 400626.00000 \\
        2 & 1014070.00000 & 13001100.00000 & 20095100.00000 \\
        3 & 8.73205 & 39.37720 & 47.23300 \\
        4 & 1.01269 & 1.49320 & 2.23439 \\
        5 & 1.04000 & 1.04009 & 1.04024 \\
        6 & 1.00255 & 1.01136 & 1.04671 \\
        7 & 1.00103 & 1.00493 & 1.07664 \\
        8 & 1.01751 & 1.07809 & 1.14315 \\
        9 & 1.02105 & 1.08912 & 1.14154 \\
        10 & 1.16098 & 2.47356 & 4.22304 \\
        11 & 1.22269 & 2.99397 & 5.27875 \\
        12 & 1.00017 & 1.00008 & 1.00016 \\
        13 & 1.00004 & 1.00033 & 1.00049 \\
        14 & 1.00011 & 1.00013 & 1.00022 \\
        15 & 1.00059 & 1.01411 & 1.05708 \\
        16 & 1.00140 & 1.00659 & 1.01079 \\ \hline
        17 & 233.67300 & 3743.47000 & 23101.20000 \\
        18 & 8.23793 & 234.69800 & 376.02300 \\
        19 & 1.00137 & 1.01701 & 1.05380 \\
        20 & 1.12523 & 10.67250 & 19.02400 \\
        21 & 2.50779 & 77.06230 & 3845.59000 \\
        22 & 1.00992 & 1.22098 & 1.70789 \\ \hline
        23 & 1.08696 & 1.08696 & 1.08696 \\
        24 & 1.05009 & 1.08333 & 1.08333 \\
        25 & 1.06362 & 1.08000 & 1.08000 \\
        26 & 1.03850 & 1.07692 & 1.07692 \\
        27 & 1.00220 & 1.07407 & 1.07407 \\
        28 & 1.07143 & 1.07143 & 1.07143 \\
        29 & 1.06897 & 1.06897 & 1.06897 \\
        30 & 1.06667 & 1.06667 & 1.06667 \\ \hline
    \end{tabular}
  \end{center}
\end{table*}

\subsubsection{Dependence on the number of function evaluations}

Again, we use $10^5$ function evaluations regardless of the dimension.

In 10 dimensions, on functions 5, 12, 13, 14, 15, 16, 19, 23, 26, 28, 29, and 30, there is almost no progress between $10^4$ and $10^5$ function evaluations. 

\begin{figure}
    \begin{center} 
    \includegraphics[width=7cm]{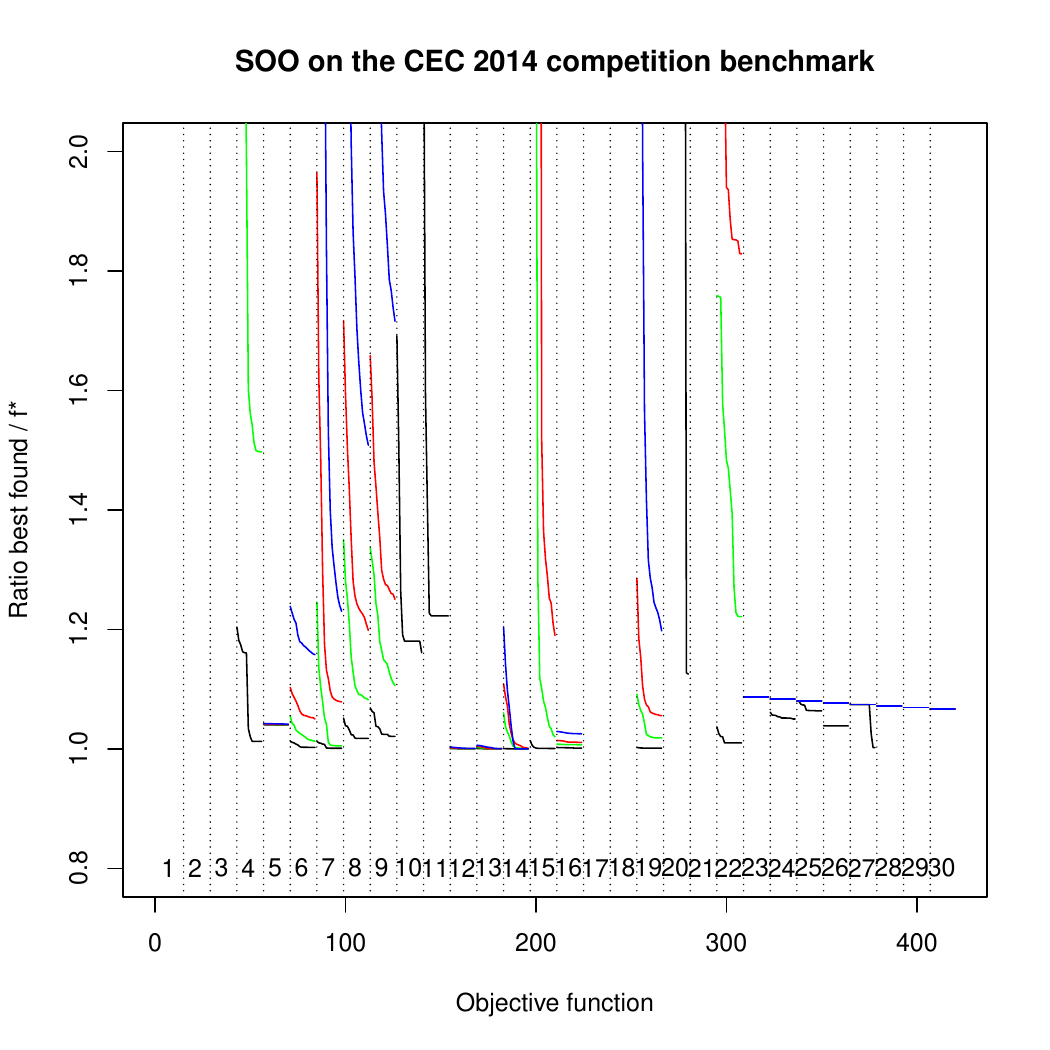}
    \caption{For each of the 30 functions of the competition, we plot the ratio between the best point found and the optimum value along the optimization. We restrict the illustration to ratios that are $\le 2$. Colors indicate the dimension of the problem: black is 10D, green is 30D, red is 50D, and blue is 100D. Naturally, the larger the dimension, the larger the ratio in general.}
  \label{fig:sooAlongIterations}
  \end{center}
\end{figure}

Figure \ref{fig:sooAlongIterations} provides a graphical illustration of the improvement along function evaluations.

\subsubsection{Comparison with DiRect}

We use the implementation of DiRect available in the NLopt library \cite{nlopt} version 2.4.1. Due to lack of time, we compared DiRect and SOO only on problems in 10 dimensions. We perform $10^5$ function evaluations at each run.

SOO outperforms DiRect on 21 functions and provides the same optimum on 5 other functions (\cf{} table \ref{table:SOOvsDiRect}). In our opinion, this is a strong case for SOO which enjoys formal properties that DiRect does not.

\begin{table*}
  \begin{center}
    \caption{Comparison of the results obtained with SOO and DiRect on the 30 functions in 10 dimensions. On each function, each algorithm performs $10^5$ evaluations. Bold face indicates the best result for each function.}
    \label{table:SOOvsDiRect}
    \begin{tabular}{ccc}
      Function & DiRect & SOO \\ \hline
        1 & 75280.30000 & \textbf{1139.36000} \\
        2 &     \textbf{3.57200} & 1014070.00000 \\
        3 &    21.44010 & \textbf{8.73205} \\
        4 &     \textbf{1.00071} & 1.01269 \\
        5 &     1.04000 & 1.04000 \\
        6 &     1.00712 & \textbf{1.00255} \\
        7 &     \textbf{1.00070} & 1.00103 \\
        8 &     1.03980 & \textbf{1.01751} \\
        9 &     1.03427 & \textbf{1.02105} \\
        10 &     1.60417 & \textbf{1.16098} \\
        11 &     2.40851 & \textbf{1.22269} \\
        12 &     1.00026 & \textbf{1.00017} \\
        13 &     1.00014 & \textbf{1.00004} \\
        14 &     1.00012 & \textbf{1.00011} \\
        15 &     1.00128 & \textbf{1.00059} \\
        16 &     1.00194 & \textbf{1.00140} \\ \hline
        17 &   330.52100 & \textbf{233.67300} \\
        18 &     \textbf{8.11702} & 8.23793 \\
        19 &     1.00208 & \textbf{1.00137} \\
        20 &     5.54139 & \textbf{1.12523} \\
        21 &    12.54650 & \textbf{2.50779} \\
        22 &     1.20054 & \textbf{1.00992} \\ \hline
        23 &     1.08696 & 1.08696 \\
        24 &     1.05581 & \textbf{1.05009} \\
        25 &     1.08000 & \textbf{1.06362} \\
        26 &     1.03859 & \textbf{1.03850} \\
        27 &     1.07407 & \textbf{1.00220} \\
        28 &     1.07143 & 1.07143 \\
        29 &     1.06897 & 1.06897 \\
        30 &     1.06667 & 1.06667 \\ \hline
    \end{tabular}
  \end{center}
\end{table*}

\subsubsection{Post-processing with a local optimizer}

As seen above, increasing the amount of function evaluations does
improve the result. However, the computational effort seems a little
bit too much with regards to the gain. Actually, SOO does not provide
a local optimum, only the best point it has found. Obviously, the
results may be post-processed by some derivative free local
optimizer. To respect the CEC'2014 competition setting, we did not
consider gradient descent algorithms which requires the gradient of
the objective function. We used derivative-free local optimizers,
namely BOBYQA \cite{bobyqa} and SBPLX \cite{subplx}, as available in
the already mentioned NLopt library.

The local optimizer also evaluates the objective function so that a trade-off between the number of evaluations by SOO and the local optimizer has to be found. There is also the issue of which point to locally optimize: indeed, SOO provides a set of points that may be optimized, not just one.

We have not yet studied thoroughly these issues, neither experimentally, nor theoretically. So, we assign 5\% of the function evaluations to the local optimization of the best point found by SOO.

Table \ref{table:postprocess} gives the results on the 30 functions in dimension 10 and 30, using the BOBYQA optimizer (results obtained with SBPLX are not as good as those obtained with BOBYQA).

In 10 dimensions, the results are now very close to the optimum for all 30 functions. In 30 dimensions, a large majority of functions are optimized very close to their optimum.

This improvement was expected; it witnesses the fact that SOO does a very good job at identifying an area in which optimal or almost optimal points lay. Getting to the optimum is then only a matter of local optimization.

\begin{table*}
  \begin{center}
    \caption{For each function of the CEC'2014 competition defined in dimensions 10, and 30, this table gives the ratio $\frac{f(x^*)}{f^*}$ obtained after post-processing the best point found by SOO with BOBYQA. In each dimension, the number of objective function evaluations is $10^4\times$dimension of the problem. This is CEC'2014 competition setup. The results in the first and third columns are the same as in table \ref{table:dim}; we mention them in the table to ease comparisons.}
    \label{table:postprocess}
    \begin{tabular}{ccccc}
      Function & \multicolumn{2}{c}{10D} & \multicolumn{2}{c}{30D} \\ 
               & no post-optimization & post-optimization
               & no post-optimization & post-optimization \\ \hline
        1 & 1139.36000 & 1.00000 & 502760.00000 & 309.30400 \\
        2 & 1014070.00000 & 1.00656 & 13003300.00000 & 1.02801 \\
        3 & 8.73205 & 1.00004 & 41.86750 & 20.82870 \\
        4 & 1.01269 & 1.01084 & 1.49726 & 1.16639 \\
        5 & 1.04000 & 1.04000 & 1.04034 & 1.04009 \\
        6 & 1.00255 & 1.00255 & 1.01285 & 1.01123 \\
        7 & 1.00103 & 1.00064 & 1.00500 & 1.00500 \\ 
        8 & 1.01751 & 1.01741 & 1.08213 & 1.07338 \\
        9 & 1.02105 & 1.01879 & 1.10622 & 1.08181 \\
        10 & 1.16098 & 1.11888 & 2.76473 & 2.19978 \\
        11 & 1.22269 & 1.21886 & 3.17396 & 3.17396 \\ 
        12 & 1.00017 & 1.00017 & 1.00015 & 1.00009 \\
        13 & 1.00004 & 1.00004 & 1.00034 & 1.00034 \\ 
        14 & 1.00011 & 1.00011 & 1.00089 & 1.00013 \\
        15 & 1.00059 & 1.00046 & 1.02031 & 1.01377 \\
        16 & 1.00140 & 1.00140 & 1.00697 & 1.00683 \\ \hline
        17 & 233.67300 & 1.07946 & 3880.91000 & 3.90322 \\
        18 & 8.23793 & 1.00286 & 1945.28000 & 2.07380 \\
        19 & 1.00137 & 1.00113 & 1.01835 & 1.00749 \\
        20 & 1.12523 & 1.00564 & 13.64210 & 1.89537 \\
        21 & 2.50779 & 1.08727 & 99.77110 & 99.77110 \\ 
        22 & 1.00992 & 1.00992 & 1.22135 & 1.22135 \\ \hline 
        23 & 1.08696 & 1.08696 & 1.08696 & 1.08696 \\
        24 & 1.05009 & 1.04934 & 1.08333 & 1.08333 \\
        25 & 1.06362 & 1.06302 & 1.08000 & 1.08000 \\
        26 & 1.03850 & 1.03850 & 1.07692 & 1.07692 \\
        27 & 1.00220 & 1.00220 & 1.07407 & 1.07407 \\
        28 & 1.07143 & 1.07143 & 1.07143 & 1.07143 \\
        29 & 1.06897 & 1.06897 & 1.06897 & 1.06897 \\
        30 & 1.06667 & 1.06667 & 1.06667 & 1.06667 \\ \hline
    \end{tabular}
  \end{center}
\end{table*}

\section{Conclusion and future work}

In this paper, we have introduced the idea that the optimization of a function available through a black-box may be formulated as a problem of sequential decision making under uncertainty. Function optimization is thus a bandit problem, the aim being to find the best action to do after a fixed known amount of function evaluations. By so doing, we inherit the rich theory of bandits. This leads us to introduce the algorithm SOO. SOO is designed in order to meet certain formal properties regarding its performance after a finite amount of function evaluations. In this paper, we provide an experimental study of SOO on the CEC2014 competition on single objective real parameter numerical optimization. Though probably not the best global optimizer available today, we think that SOO has demonstrated its interest as a global optimization algorithm. In particular, against common intuition, and against our own expectation, SOO is able to perform remarkably in high dimension functions, the degradation of performance 
between 10D and 100D being sometimes very small. SOO is a very simple algorithm, hence very simple to implement, and rather fast. Moreover, SOO may be combined with a local optimization algorithm, then providing greatly improved results.

After this work, there are many questions that we are going to study further. 1)
Regarding parameter tuning, we have not done an extensive study, and we think
that some parameters may take advantage of an online auto-tuning. 2) For lack of
time, we have not tried to compete on the part B of the competition in which the
amount of function evaluations is very small, but one may learn a model to find
a better optimum, or even guide the search. There are very interested
theoretical questions behind that. 3) Being a global optimizer, SOO is not at
ease with local optimization in high dimensions; so, it makes sense to post-process the best
point(s) found by SOO by some local optimization procedure. We did some
experiments using this idea, though a more complete experimental study is due;
furthermore, again, there are very interesting questions to investigate from a
theoretical point of view, in particular the trade-off between the SOO and the
local optimizer. 4) The characterization of a cell by the point in its center
may also be a
path of future work. Restricting the behavior of the objective function by its value on a single point of a cell, and particularly its center, is really a choice that should be challenged. We may consider sampling different points of the cell to determine some sort of average behavior in the cell, and an associated higher order moments. From this, we would be able to derive a confidence bound regarding each cell that would guide the choice of the next cell to split. This links this work with its original motivation of optimizing non deterministic functions. 
Indeed, SOO is one member of a family of algorithms, all based on the
same principles. Other algorithms assume certain properties on the
objective function (HOO and DOO assumes that the local smoothness of the
function is known and is required in tuning the algorithm); StoSOO \cite{icml2013b} is
an extension of SOO which performs
function optimization of noisy (or stochastic) function. The
interested reader may find useful R.\@ Munos' survey
\cite{MunosFTML2013}. All these algorithms may optimize functions
having various properties; this approach may lead to other algorithms.


\bibliography{bandits-attack-fct-opt}
\bibliographystyle{plain}

\end{document}